\documentclass[journal]{IEEEtran}
\usepackage[utf8]{inputenc}
\usepackage{graphicx}

\makeatletter
\usepackage{cite}

\@ifundefined{showcaptionsetup}{}{%
 \PassOptionsToPackage{caption=false}{subfig}}
\usepackage{subfig}
\makeatother

\begin{document}
\title{Spatial-Temporal Generative AI for Traffic Flow Estimation with Sparse
Data of Connected Vehicles}
\author{Jianzhe~Xue, Yunting~Xu, Dongcheng~Yuan, Caoyi~Zha, Hongyang Du,
Haibo~Zhou, and Dusit~Niyato,~\IEEEmembership{Fellow,~IEEE}\\
 % <-this % stops a space
\IEEEcompsocitemizethanks{ \IEEEcompsocthanksitem Jianzhe Xue,
Yunting Xu, Dongcheng Yuan, Caoyi Zha, and Haibo Zhou are with the
School of Electronic Science and Engineering, Nanjing University,
Nanjing 210023, China. E-mail: \{jianzhexue, yuntingxu, dongchengyuan\}@smail.nju.edu.cn;
zhacaoyi@163.com; haibozhou@nju.edu.cn. \IEEEcompsocthanksitem Hongyang
Du and Dusit~Niyato is with the School of Computer Science and Engineering,
Nanyang Technological University, Singapore. Email: hongyang001@e.ntu.edu.sg;
dniyato@ntu.edu.sg. }}
\maketitle
\begin{abstract}
Traffic flow estimation (TFE) is crucial for intelligent transportation
systems. Traditional TFE methods rely on extensive road sensor networks
and typically incur significant costs. Sparse mobile crowdsensing
enables a cost-effective alternative by utilizing sparsely distributed
probe vehicle data (PVD) provided by connected vehicles. However,
as pointed out by the central limit theorem, the sparsification of
PVD leads to the degradation of TFE accuracy. In response, this paper
introduces a novel and cost-effective TFE framework that leverages
sparse PVD and improves accuracy by applying the spatial-temporal
generative artificial intelligence (GAI) framework. Within this framework,
the conditional encoder mines spatial-temporal correlations in the
initial TFE results derived from averaging vehicle speeds of each
region, and the generative decoder generates high-quality and accurate
TFE outputs. Additionally, the design of the spatial-temporal neural
network is discussed, which is the backbone of the conditional encoder
for effectively capturing spatial-temporal correlations. The effectiveness
of the proposed TFE approach is demonstrated through evaluations based
on real-world connected vehicle data. The experimental results affirm
the feasibility of our sparse PVD-based TFE framework and highlight
the significant role of the spatial-temporal GAI framework in enhancing
the accuracy of TFE.
\end{abstract}


\section{Introduction}

Traffic flow estimation (TFE) plays a crucial role in intelligent
transportation systems (ITS), providing real-time traffic flow information
to ITS applications including route planning and traffic management
\cite{TFP_Network,TSE_Highway}. Mobile crowdsensing (MCS) with connected
vehicles offers a way for TFE that uses the average speeds of probe
vehicles in each region or road segment to represent traffic flow
\cite{GraphSAGE,URLLC}. Given the cost of data collection and user
privacy, obtaining a massive amount of probe vehicle data (PVD) is
difficult in practical applications. As a result, TFE typically needs
to be performed on a sparse PVD obtained from a limited number of
sparsely distributed connected vehicles throughout the city.

However, inadequate PVD poses a challenge to accurately estimating
traffic flow \cite{SMC_survey}. According to the central limit theorem
in statistics, the mean of a small number of samples has a Gaussian
error relative to the mean of the entire population. For TFE using
sparse PVD, data sparsification results in fewer samples in each region,
which further introduces Gaussian errors into the estimates of each
region. In addition, the PVD sparsification sometimes even results
in the absence of samples in certain areas. Therefore, the initial
estimation that directly uses the average speeds obtained from sparse
PVD cannot provide an accurate estimation. The initial estimation
by sparse PVD is subject to errors and misses in the initial estimations
compared to the ideal estimation using average speeds obtained from
a massive amount of PVD \cite{GraphSAGE}. 

Fortunately, the presence of spatial-temporal correlations in the
traffic flow makes it possible to further improve the accuracy of
the initial estimations \cite{TGASA}. Spatial-temporal correlations
reflect the inherent continuity and interconnectedness of traffic
movements over time and space. Changes in traffic flow are not isolated
to specific moments or locations but unfold continuously, demonstrating
a seamless progression across both temporal and spatial dimensions
\cite{NC_Traffic_Jams,TGCN}. Temporal correlations in traffic flow
may be influenced by various factors, while spatial correlations represent
how traffic conditions vary across different geographical regions.
With the information provided by the spatial-temporal correlation,
we can modify the error or interpolate the missing with generative
artificial intelligence (GAI) techniques by generating new TFE outcomes.

Conditional GAI is a powerful technique for generating high-quality
and accurate TFE outcomes from low-quality initial estimations, contributing
to cost-effective TFE from sparse PVD \cite{CurbGAN,FillMissing_Niyato}.
Specifically, we propose to design a conditional GAI framework for
TFE consists of two parts, the conditional encoder and the generative
decoder. The conditional encoder is a spatial-temporal deep neural
network for mining spatial-temporal correlations in the initial TFE
results \cite{STGAISurvey}. The output of the conditional encoder
is a latent representation of the extracted spatial-temporal correlations,
which will be used as conditions for the decoder. Based on the given
conditions, the generative decoder utilize the generative model to
produce new high-quality and accurate TFE outcomes as final outcomes
\cite{AIGC_Niyato}. The utilization of conditional GAI can effectively
overcome the problem of low-quality estimation results caused by data
sparsification and improve the accuracy of TFE.

\begin{figure*}
\centering \includegraphics[width=7in]{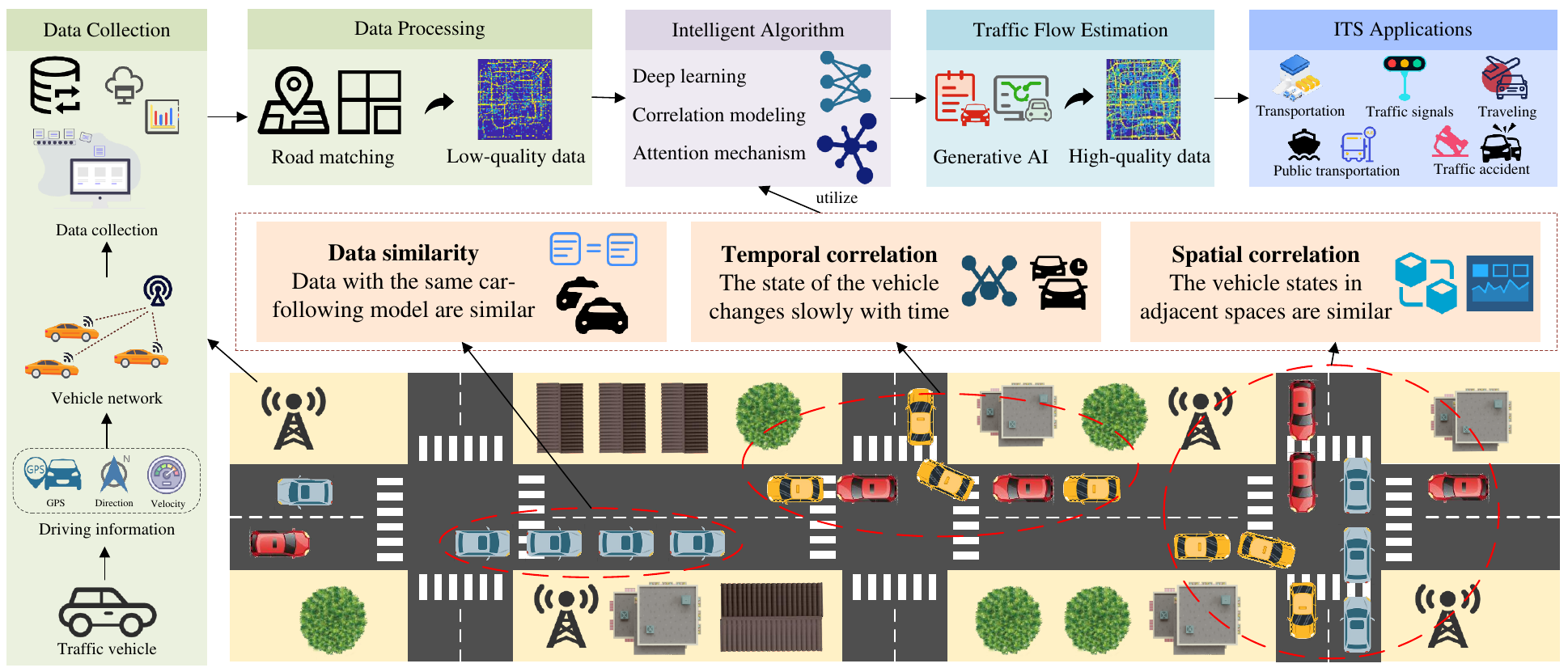}
\caption{The cost-effective sparse mobile crowdsensing traffic flow estimation
framework.}
\label{fig:Framework}
\end{figure*}

In this paper, we propose a cost-effective sparse MCS approach for
real-time TFE, which utilizes conditional GAI to generate accurate
estimation using only the limited PVD. Specifically, we first collect
sparse PVD, which includes GPS and travel speeds of connected vehicles
that are sparsely and uniformly distributed. Then, the initial estimations
are obtained by directly using the average speed of probe vehicles
in each region or road segment. Finally, we employ spatial-temporal
conditional GAI to refine and improve TFE outcomes, transforming the
low-quality initial TFE results into accurate estimations. The major
contributions of this paper are delineated as follows:
\begin{itemize}
\item We propose a cost-effective sparse MCS approach for real-time TFE,
where the data collocation overhead is reduced by collecting sparse
PVD. The impact of data sparsification on TFE accuracy is analyzed.
\item We design a spatial-temporal conditional GAI framework to improve
the TFE accuracy, which consists of a conditional encoder and a generative
decoder. We further discuss spatial-temporal neural networks used
in encoder for traffic correlations mining, and generative models
used in decoder for generating accurate TFE outcomes. 
\item Through a case study using a real dataset, we demonstrate the feasibility
and validity of spatial-temporal conditional GAI for cost-effective
TFE, which can be accurately estimated using sparse PVD.
\end{itemize}
\author{\\}

The remainder of this paper is organized as follows. Section II introduces
the TFE using sparse PVD. Section III discusses the challenges in
accurate TFE using sparse PVD. Section IV presents the spatial-temporal
conditional GAI framework for TFE. Section V shows the case study
of GAI-enabled TFE using sparse PVD. Section VI concludes this paper.

\section{Sparse Mobile Crowdsensing for Traffic State Estimation}

TFE has multiple applications in ITS. Firstly, TFE underpins real-time
traffic monitoring and management, empowering transportation authorities
to rapidly address congestion, accidents, and various incidents. Secondly,
TFE is instrumental in traffic planning and the optimization of infrastructure.
By analyzing traffic flow patterns and trends, planners are equipped
to formulate more strategic transportation plans to enhance the overall
efficiency of traffic systems. Thirdly, TFE allows for the dynamic
adjustment of traffic light timings in sync with real-time traffic
conditions, which minimizes wait times and optimizes traffic flow
at intersections.

TFE typically employs roadside sensors or connected vehicles to collect
and analyze real-time vehicle movement situations \cite{TSE_GPS,RAN_survey}.
TFE via roadside sensors requires a large number of devices to be
deployed, which is costly. In contrast, the use of sparse PVD is a
more cost-effective alternative to traditional methods. Due to the
inherent spatial and temporal correlation of traffic flows, PVD exhibits
a certain degree of redundancy. For example, vehicles nearby will
have similar speeds. This redundancy amplifies the burden of data
collection. To address these issues, the innovative sparse MCS has
been introduced. Sparse MCS effectively diminishes the overhead of
PVD acquisition while preserving the integrity and precision of the
estimation, capitalizing on spatial-temporal traffic correlations.

\subsection{Framework of Sparse MCS}

Sparse MCS is an approach within the broader concept of MCS that emphasizes
the collection of data from a selected set of devices or areas, rather
than the continuous collection of data from all sources \cite{SMC_survey}.
Reducing data requirements helps to minimize waste and overhead of
data collection, transmission, and storage resources, but additional
steps need to be taken to ensure that sensing accuracy can meet requirements. 

Our sparse MCS TFE approach is shown in Fig. \ref{fig:Framework},
which improves the estimation accuracy through conditional GAI based
on spatial-temporal correlation in traffic flow. The first step is
to collect driving information through vehicular networks, mainly
including vehicle GPS coordinates, vehicle traveling direction, and
vehicle speed. In sparse MCS, the sparse PVD is collected from sparsely
and uniformly distributed connected vehicles across the entire urban
area. Unlike sparing the data by region or road segment, we sparse
the data from the source for the following reasons. Firstly, the high
mobility of vehicles complicates the demarcation of regions. Second,
it is impractical to collect information of all vehicles within the
road network, and therefore, available PVD can be considered to be
sparsed at the data source. Thirdly, collecting PVD from widely distributed
vehicles provides a more comprehensive coverage and hence a more complete
picture of the traffic situation.

Once the sparse PVD is collected, the next step is data processing.
Road matching is first performed, which involves comparing the vehicle's
GPS coordinates with known road maps to accurately determine the specific
road or region on which the vehicle is traveling. Subsequently, we
calculated low-quality initial TFE results by calculating the average
speed of the vehicles in each region over a specified past time. These
average speeds represent the current traffic flow speed in each region.
Initial TFE results from previous and current times are used as inputs
to the spatial-temporal conditional GAI. The GAI uses the conditional
encoder to mine spatial-temporal correlations in the initial TFE results
and uses the generative decoder to generate new high-quality and accurate
TFE outcomes as final outcomes.

\subsection{Benefit of Sparse MCS}

The application of sparse MCS has significant advantages in terms
of cost-effectiveness, scalability, and robustness:

\subsubsection{Cost-effectiveness}

Sparse MCS is cost-effective by collecting smaller quantities of PVD.
This reduces unnecessary expenses associated with the collection of
large volumes of PVD, including overhead associated with acquisition,
transmission, and storage. It fully optimizes the use of available
resources and maximizes the value of the data. 

\subsubsection{Scalability}

Unlike traditional road sensors, which are limited by their physical
location and coverage, sparse MCS can take advantage of the wide coverage
of connected vehicles. As long as the vehicles are equipped with the
appropriate sensors, the sensing network has the potential to extend
to any area, providing a highly scalable solution for a variety of
applications and environments. 

\subsubsection{Robustness}

Sparse MCS is robust and reduces the stringent requirements on PVD
quality. This approach is robust enough to cope with situations where
PVDs may be scarce or missing in certain regions or at specific times,
ensuring that the system remains effective despite reduced data volumes.
For startups, this robustness allows them to achieve accurate TFE
from relatively small amounts of user data, providing precise information
that is invaluable for other applications.

\section{Challenges of Accurate TFE from Sparse PVD}

In general, the accuracy of MCS is highly dependent on the amount
of data collected, as is the case for TFE. The accuracy is also reduced
when using sparse PVD to directly estimate traffic flow. Fortunately,
by exploiting the spatial-temporal correlation of traffic flow, the
accuracy of TFE can be further improved.

\begin{figure}
\centering \includegraphics[width=3.4in]{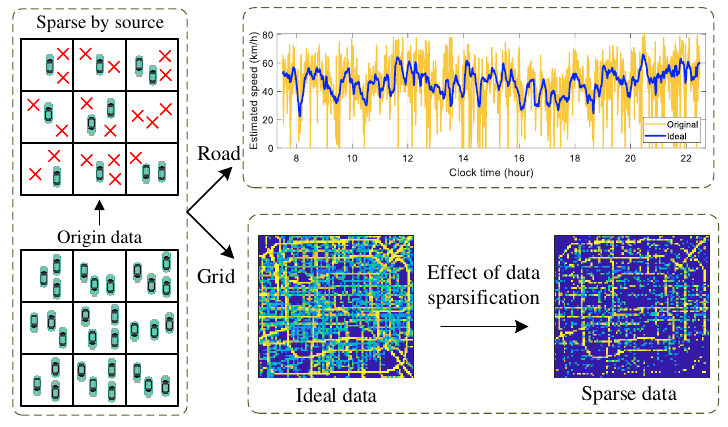}
\caption{The impact of data sparsification.}
\label{fig:TFE}
\end{figure}

\begin{figure}
\centering \includegraphics[width=3.4in]{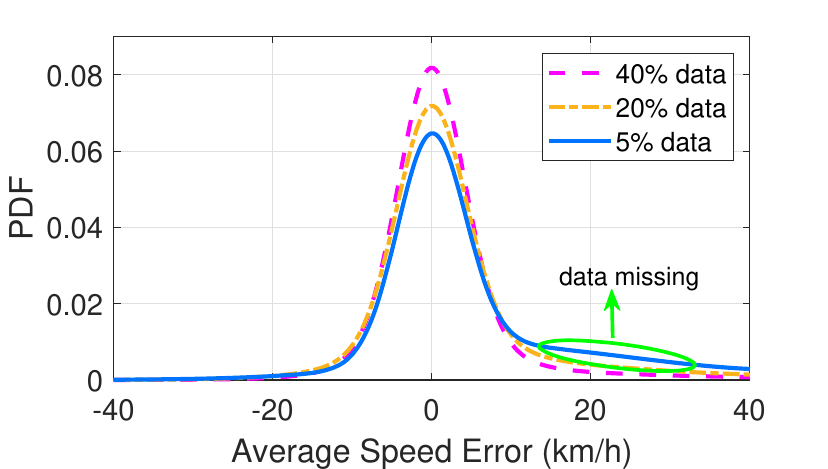} \caption{The impact of data sparsification.}
\label{fig:Error}
\end{figure}

\subsection{Effect of PVD Sparsification}

Data sparsification for PVD has two significant effects on the initial
TFE results, namely accuracy and completeness \cite{TGASA}. The reduced
sample size associated with data sparsification affects the accuracy
of the estimates. This is because averages derived from a limited
number of samples may be biased or inaccurate. Besides, the sparse
PVD collected may provide incomplete coverage of roads, resulting
in missing estimates for some roads.

As shown in Fig. \ref{fig:TFE}, the PVD for TFE is sparsed at the
data source, i.e., information is collected from fewer connected vehicles.
For a single road, the blue curve represents an ideal, high-quality
estimations of traffic flow derived from a large number of PVD, and
the yellow curve represents the estimation derived from sparse PVD.
Zero values in the yellow curve indicate missing estimates due to
the lack of collected PVD for the road at a given time. The speeds
estimated from the sparse PVD tend to fluctuate around the ideal value
and occasionally have missing data points. For a city-wide estimation,
the left inset shows a map of ideal traffic flow speeds derived from
a large number of PVD, while the right inset shows a map of speeds
derived from sparse PVD. The above results demonstrate that data sparsification
introduces errors and missing values in the initial TFE results. 

Fig. \ref{fig:Error} shows the distribution of city-wide estimation
errors for different levels of data sparsity. These errors typically
follow a Gaussian distribution, consistent with the Central Limit
Theorem, which states that the mean of a small number of samples will
exhibit a Gaussian error relative to the mean of the entire population.
Notably, the positive portion of the curve is higher than the negative
portion. This asymmetry in the distribution of the error is caused
by missing data since the missing values set to zero therein are always
smaller than the ideal value. Therefore, it is possible to make rough
estimations of traffic flow with sparse data despite the presence
of Gaussian errors and missing values.

\subsection{Spatial-Temporal Correlations}

Spatial-temporal correlations of traffic flows are formed by complex
dependencies of traffic data in the spatial and temporal dimensions.
In transport systems, the state of traffic at one location affects
or is correlated with the state of traffic at another location, and
these relationships change over time. Spatial correlation means that
traffic behavior in one region will affect nearby regions, often leading
to congestion spreading throughout the road network, e.g. congestion
on one road will likely lead to congestion on neighboring roads. Temporal
correlation describes patterns and relationships in the evolution
of traffic conditions over time and is characterized by cyclical changes
in traffic flows with continuous fluctuations rather than sudden changes.

\subsection{Key Challenges}

The use of sparse PVD for accurate TFE faces significant challenges
due to the effects of PVD sparsification. Thankfully, the use of spatial-temporal
correlations in traffic flow can lead to accurate estimations from
initial estimations. With the additional information provided by spatial-temporal
correlations, we can fully understand the dynamic properties of the
traffic network. Even in sparse PVD, there are still rich spatial-temporal
correlations that help to accurately reconstruct the traffic flows
and thus achieve accurate TFE. However, achieving accurate and comprehensive
traffic flow estimates using sparse PVD remains a formidable challenge
\cite{ST_DataMining}. This difficulty mainly arises from the complexity
involved in capturing and effectively utilizing spatial-temporal correlations
in dynamic traffic flows, which are inherently intricate.

\section{Spatial-Temporal Conditional GAI}

\begin{figure*}
\centering \includegraphics[width=7in]{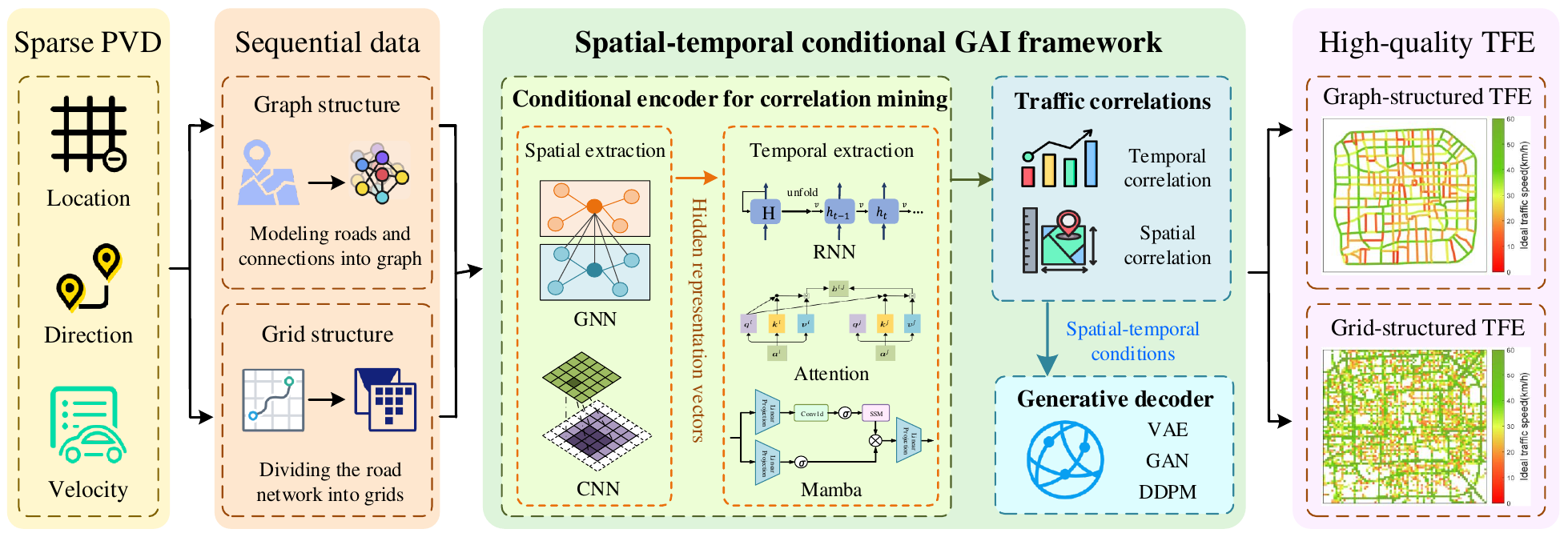}
\caption{An illustration of conditional GAI framework for traffic flow estimation.}
\label{fig:GAI} 
\end{figure*}

The spatial-temporal conditional GAI framework, shown in Fig. \ref{fig:GAI},
employs an encoding-decoding architecture to generate high-quality
outcomes and enhance the accuracy of TFE by mining spatial-temporal
correlations in traffic data. Depending on the road map format, these
initial estimations obtained from sparse PVD are organized as either
grid-structured or graph-structured data. By arranging these estimations
chronologically including both historical and current data, we create
a sequence of grid or graph-structured estimations with rich spatial-temporal
correlations. These sequential data are used as inputs to the conditional
encoder, which extracts and utilizes these correlations as conditions
for the generative decoder. Subsequently, the generative decoder uses
these conditioned inputs to produce high-quality, accurate TFE outcomes. 

The cornerstone of our proposed framework is the effective capture
of spatial-temporal correlations in traffic flow, achieved through
the conditional encoder. The backbone of the conditional encoder is
the spatial-temporal deep neural network, which consists of two modules
cascaded together, one responsible for capturing spatial correlation
and the other for capturing temporal correlation. The following offers
an overview of the various neural network models that can be employed
within the spatial-temporal conditional encoder, highlighting their
capabilities and roles in enhancing the accuracy of TFE.

\subsection{Models for Spatial Correlations Mining}

Spatial correlations delineate the intricate interconnections among
various regions within a transportation network. To efficaciously
extract the spatial correlations embedded in data formats structured
as grids or graphs, the deployment of specialized deep neural networks
is indispensable, adapting the complexities inherent in traffic flow
data. The judicious selection and application of optimal neural network
architectures are crucial for accurately capturing these spatial correlations,
thereby significantly enhancing the accuracy of TFE.

\subsubsection{Grid}

Grid data is arranged in a matrix-like grid pattern, with each value
occupying a unique position across rows and columns. This structured
approach is leveraged to partition urban maps into uniform grids,
resulting in grid-based representations of traffic dynamics. Two primary
models can be used to learn spatial correlations of grid data: the
convolutional neural network (CNN) and the vision transformer (ViT).
The CNN discerns local features and spatial patterns via convolutional
and pooling layers. It is particularly proficient in identifying localized
traffic correlations, offering granular insights into traffic conduct.
Conversely, the ViT segments images into patches, employing self-attention
mechanisms to uncover spatial interrelations among these patches.
It is adept at delineating the complex spatial interplay characteristic
of traffic flow, thus providing a comprehensive view of traffic dynamics
over expansive areas.

\subsubsection{Graph}

Graph data structures encapsulate information through a network of
nodes and edges, where nodes typically denote road segments, and edges
signify the topological links between them. It transforms city maps
into graphs with roads as the main focus, associating traffic flow
data with road segments. Graph neural networks (GNNs) are adept at
managing such graph-structured data. The graph convolutional network
(GCN) employs a localized first-order approximation of spectral graph
convolutions to aggregate information within the graph. Enhancing
this methodology, the graph attention network (GAT) assigns varying
attention weights to nodes, emphasizing significant node characteristics
and their connections. Besides, the graph sample and aggregate neural
network (GraphSAGE) is designed to learn a function that generates
embeddings by sampling and aggregating features from a node’s local
neighborhood. 

\subsection{Models for Temporal Correlations Mining}

Temporal correlation in traffic flow encapsulates the interdependencies
between traffic conditions across different time intervals. It suggests
that present traffic states are not isolated events but are consequentially
linked to antecedent and subsequent traffic states. To adeptly harness
these correlations, it is imperative to meticulously process sequential
data and engage neural networks that are expressly designed to capture
temporal dynamics. 

\begin{figure*}[!t]
\centering\subfloat[The CRNet-based GAI framework for improving TFE accuracy on grid-structured
data.]{\includegraphics[width=7in]{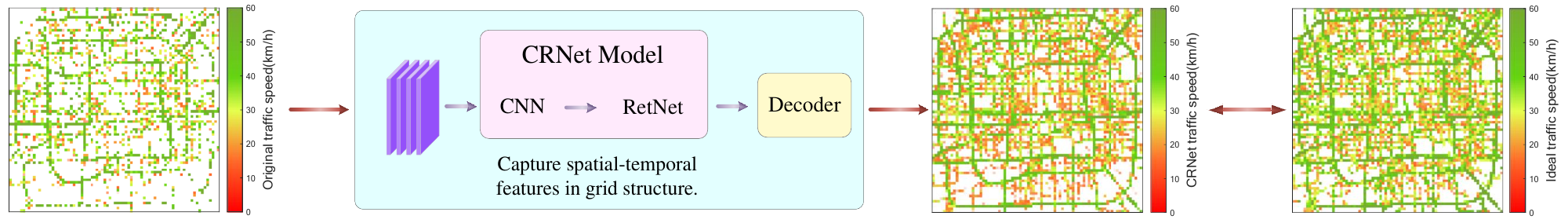}\label{fig:process-CRNet}

}

\subfloat[The TGASA-based GAI framework for improving TFE accuracy on graph-structured
data.]{\includegraphics[width=7in]{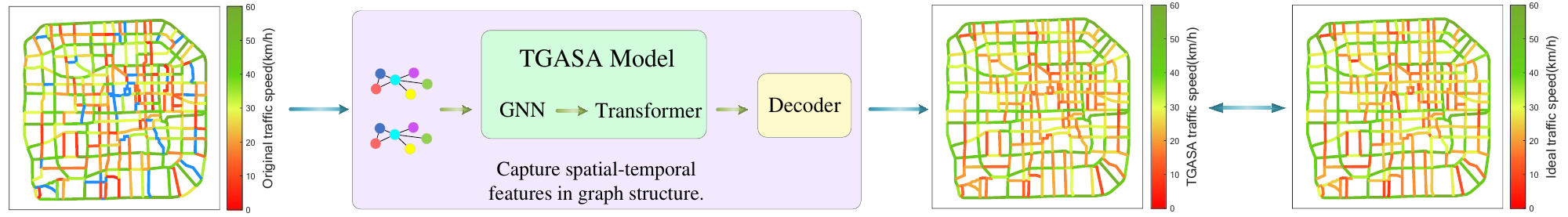}\label{fig:process-TGASA}

}

\caption{The visualization of TFE accuracy improvement using generative AI.}
\label{fig:process}
\end{figure*}

\subsubsection{RNN}

Recurrent neural networks (RNNs) are distinguished by their cyclical
connections, enabling them to preserve a memory of antecedent inputs.
The archetypal RNN iteratively updates its latent state at each temporal
step by integrating the current input with the preceding latent state,
facilitating the extraction of patterns and the retention of information
across the data sequence. To overcome the difficulty of vanishing
or exploding gradients, the advent of long short-term memory (LSTM)
networks introduces specialized memory cells and gating mechanisms,
empowering LSTMs to judiciously regulate information over protracted
intervals. To further reduce the complex, the gated recurrent unit
(GRU) reduces the complexity of the model by combining input and forget
gates into a unified update gate. 

\subsubsection{Attention}

Attention-based models utilize the attention mechanism to effectively
reveal intricate temporal associations in data sequences. The transformer
with multi-head attention is adept at comprehensively modeling sequence
data and proficiently managing long-distance dependencies. The informer
uses the ProbSparse self-attention and multi-scale processing to efficiently
predict over extensive periods. The retentive network (RetNet) incorporates
a retention mechanism for sequence modeling, which supports three
computation paradigms: parallel, recurrent, and chunkwise recurrent,
aiming to achieve training parallelism, low-cost inference, and high
performance. These architectures have sophisticated attention mechanisms
that provide a dual analytical perspective including global and local
views, enabling a more nuanced understanding of traffic dynamics patterns. 

\subsubsection{State Space Model }

By integrating the state space model (SSM) architecture, Mamba synergistically
leverages the strengths of RNNs and CNNs to provide a powerful solution
for sequence data analysis. In the training phase, Mamba uses a convolutional
strategy to collectively process the input sequences to ensure efficient
operation. In the inference phase, Mamba switches to recurrent mode.
This innovative dual-mode architecture enables Mamba to capitalize
on the parallel processing prowess of CNNs while preserving the sequential
integrity inherent in RNNs. Moreover, Mamba integrates a discerning
feature selection mechanism within the SSM framework, prioritizing
salient input features and disregarding extraneous data. 

\subsection{Models for Spatial-Temporal Correlations Mining}

Capturing both spatial and temporal correlations is essential for
the proposed GAI framework to achieve accurate TFE. The neural network
models for sequential grid structure includes convolutional LSTM (ConvLSTM),
simple video prediction model (SimVP), and eidetic 3D LSTM (E3DLSTM).
As for sequential graph structures, the temporal graph convolutional
network (T-GCN), spatial-temporal graph convolutional network (STGCN),
and transformer-graph attentional sample and aggregate Neural Network
(TGASA) can be used. 

\subsubsection{Sequential Grid}

Sequential grid data refers to a group of grid-structured data arranged
in a sequential manner. ConvLSTM leverages the strengths of both CNN
and LSTM, adeptly capturing spatial features via convolutional operations
and managing temporal sequences with LSTM architectures. SimVP employs
deformable convolutions and attention mechanisms. E3DLSTM, designed
for video classification and behavior recognition, leverages a blend
of 2D and 3D convolutions. Since traffic flow data in grid format
resembles the structure of video data, these networks are also adept
at processing gridded traffic flow data.

\subsubsection{Sequential Graph }

Sequential graph data refers to a group of graph-structured data arranged
in a sequential manner. T-GCN uses both GRU and GCN to capture the
spatial-temporal dynamics of traffic patterns. STGCN uses gated sequential
convolution and spatial graph convolution to capture the relationships
among nodes in spatial-temporal graphs. TGASA combines an enhanced
GraphSAGE model with the transformer to capture the spatial correlation
and a transformer model to capture the temporal correlation. These
models employ GNN to accurately delineate intricate features of sequential
graph data.

\subsection{Generative Decoder with Spatial-Temporal Conditions}

After the spatial-temporal correlations in traffic data is encoded
into a compressed latent representation, the generative decoder uses
this information to generate TFE outcomes that resembles the ideal
estimations. It focuses on reconstructing outputs based on latent
representation conditions given by the conditional encoder. In training,
the decoder learns to minimize the difference between its outputs
and the ideal estimations. It can effectively interpolate missing
data and correct errors in the initial estimation, leading to more
accurate and high-quality estimations.

\subsubsection{Variational Auto-Encoders (VAE)}

The VAE typically consists of two parts: an encoder and a decoder.
The VAE's encoder compresses the inputs into the latent space, learning
to represent the input data in compressed form. The goal of VAE's
decoder is to have the reconstructed data be as close as possible
to the original input data, thus ensuring that the latent space representation
maintains critical information about the input data. For the application
of TFE recovery, the VAE's decoder generates accurate TFE outcomes
based on latent space conditions with spatial-temporal correlations.

\subsubsection{Generative Adversarial Networks (GAN)}

The GAN consists of a generator and a discriminator. Specifically,
the generator learns to produce output from a given input that is
indistinguishable from the real data, while the discriminator acts
as a judge and attempts to accurately distinguish between the real
and fake. The training of GANs involves alternating between optimizing
the discriminator and the generator, where the generator updates its
weights to produce samples that are harder to classify as fake. A
well-trained generator of GAN take the latent space conditions as
inputs to produce accurate TFE outcomes.

\subsubsection{Denoising Diffusion Model}

Denoising diffusion probabilistic models (DDPMs) operate by gradually
transforming data from a simple distribution into a complex distribution
through a series of small, reversible steps. This process is modeled
in two phases: the forward diffusion phase gradually converts a sample
from the data distribution into random noise through a sequence of
steps, and the reverse denoising process aims to learn how to revert
the noisy data back to its original form. The denoising process of
DDPM can be used to obtain accurate TFE outcomes with the guidance
of conditions.

\subsection{Lessons Learned}

Based on the above discussion, we present our insights for designing
the spatial-temporal conditional GAI framework for TFE from sparse
PVD. The conditional encoder firstly use the inaccurate and low-quality
initial estimations as inputs, and its the spatial module processes
the spatial features of the traffic flow at discrete moments to encode
these features into latent representations. Then, the temporal module
receives the chronologically ordered sequence of these latent representations,
identifies and integrates the temporal correlations among them. The
output of conditional encoder is a comprehensive vector that embodies
both spatial and temporal correlations, serving as the conditions
for the generative decoder. By using the generative model approach,
the generative decoder produce the accurate and high-quality TFE outcomes
based on the given conditions. 

\section{Case Study}

\subsection{Case Study Settings}

The case study utilizes the real data sourced from the Fourth Ring
Road in Beijing, handling the urban area through two distinct perspectives:
grid structure and graph structure. The resolution of the grid-structured
map is $80\times80$, while the graph road map has 500 road segments
as nodes. Our dataset includes 6 days in 2012 \cite{TGASA}. The time
period spans from 7:30 AM to 10:30 PM, and the time step is 1 minute.
The sparse data is generated by random sampling of the vehicle driving
information from the complete PVD dataset.

To show the feasibility of TFE from sparse PVD, we verify the proposed
spatial-temporal conditional GAI framework with two different approaches
based on VAE. For the convolutional retentive network (CRNet), the
encoder block utilizes CNN to aggregate spatial correlation and RetNet
to capture temporal correlation. Besides, the encoder of the TGASA
mainly comprises GNN and transformer. Both methods are meticulously
designed to capture spatio-temporal correlations inherent in traffic
data, ensuring robust and accurate TFE despite data sparsity. The
experiments are facilitated by the Tesla V100-DGXS-32GB GPU.

\subsection{Results Analysis}

\begin{figure}[!t]
\centering\subfloat[Error comparison of CRNet.]{\includegraphics[width=1.6in]{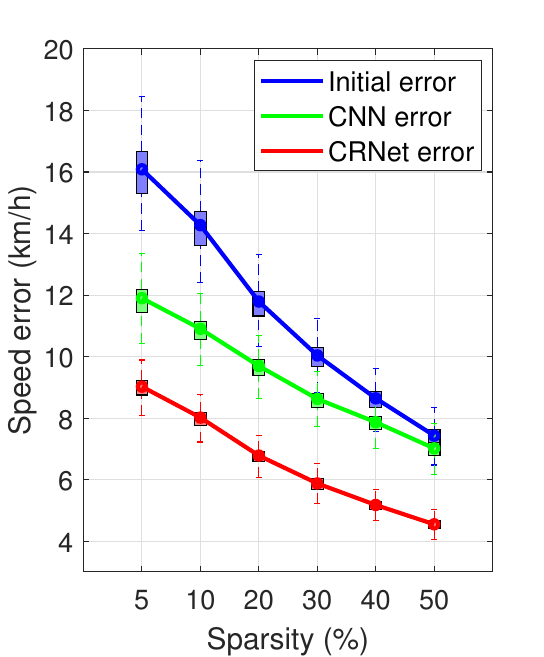}\label{fig:box-CRNet}

}\subfloat[Error comparison of TGASA.]{\includegraphics[width=1.6in]{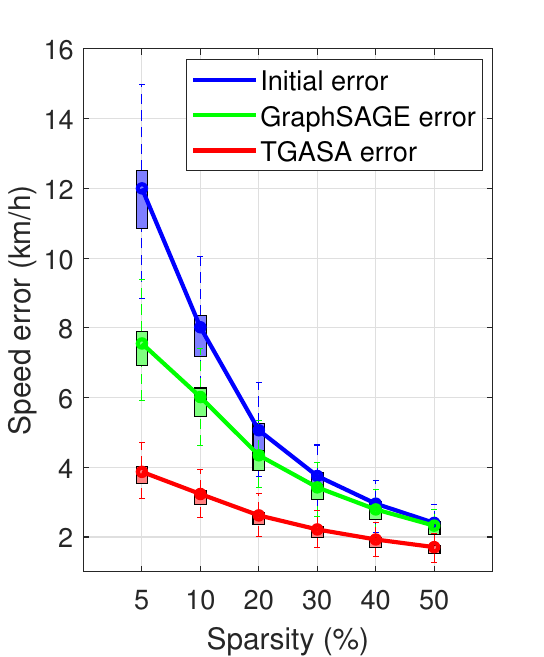}\label{fig:box-TGASA}

}\caption{The RMSE of initial and recovered.}
\label{fig:box-plot}
\end{figure}

Fig. \ref{fig:process} shows TFE recovery performance of the proposed
GAI approaches. Fig. \ref{fig:box-CRNet} shows the TFE recovery on
a grid structure, where data-absent grids are marked in white. The
initial traffic flow image derived from sparse PVD is markedly incomplete
and distorted, reflecting significant losses and inaccuracies in the
depiction of traffic speeds. After recovery, the TFE outcomes obtained
by CRNet demonstrate substantial improvement, closely resembling the
ideal estimation obtained from dense data. Fig. \ref{fig:box-TGASA}
illustrates the TFE recovery on graph structure. The road segments
where no data were collected are colored in pure blue. In the initial
TFE obtained from sparse PVD, numerous road segments appear in blue,
and the overall color distribution differs obviously from that of
ideal estimation derived from dense data. The application of TGASA
to the recovery process effectively eradicates the blue segments,
yielding a TFE that closely emulates the ideal estimation in the color
portrayal of each road segment.

Fig. \ref{fig:box-plot} shows the impact of data sparsification on
TFE accuracy and the improvement of proposed GAI approaches. The box
plots depict the distribution of root mean square errors (RMSE), while
lines represent the average RMSE across various levels of data sparsity.
As data becomes sparser, both the median and range of estimation errors
increase, along with a rise in the mean error values. Specifically,
for the case of 5\% data sparsity, the CRNet reduces the RMSE from
16.09 km/h of initial estimation to 9.02 km/h, while the TGASA reduces
the RMSE from 12.01 km/h of initial estimation to 3.87 km/h. Overall,
the proposed spatial-temporal GAI framework with the designed neural
network demonstrates substantial effectiveness in error recovery,
showing marked improvements especially as data becomes sparser. These
results underscore the necessity and feasibility of employing GAI
techniques for TFE using sparse probe vehicle data, highlighting their
potential to enhance the TFE accuracy.

\section{Conclusion}

In this paper, we have proposed a cost-effective sparse MCS approach
for TFE. To mitigate the impact of data sparsification, our approach
employs the spatial-temporal conditional GAI framework to generate
accurate and high-quality TFE outcomes. We have evaluated the feasibility
and effectiveness of the proposed TFE approach by using real PVD.
Experimental results show that our GAI approach can significantly
improve the quality of TFE outcomes, despite the initial accuracy
loss due to data sparsification. For future work, we will explore
the use of the large language models to further improve the accuracy
of TFE. \bibliographystyle{IEEEtran}
\bibliography{TFE_Mag_Ref.bib}

\begin{thebibliography}{10}
\providecommand{\url}[1]{#1}
\csname url@samestyle\endcsname
\providecommand{\newblock}{\relax}
\providecommand{\bibinfo}[2]{#2}
\providecommand{\BIBentrySTDinterwordspacing}{\spaceskip=0pt\relax}
\providecommand{\BIBentryALTinterwordstretchfactor}{4}
\providecommand{\BIBentryALTinterwordspacing}{\spaceskip=\fontdimen2\font plus
\BIBentryALTinterwordstretchfactor\fontdimen3\font minus
  \fontdimen4\font\relax}
\providecommand{\BIBforeignlanguage}[2]{{%
\expandafter\ifx\csname l@#1\endcsname\relax
\typeout{** WARNING: IEEEtran.bst: No hyphenation pattern has been}%
\typeout{** loaded for the language `#1'. Using the pattern for}%
\typeout{** the default language instead.}%
\else
\language=\csname l@#1\endcsname
\fi
#2}}
\providecommand{\BIBdecl}{\relax}
\BIBdecl

\bibitem{TFP_Network}
P.~Sun, N.~Aljeri, and A.~Boukerche, ``Machine learning-based models for
  real-time traffic flow prediction in vehicular networks,'' \emph{IEEE
  Network}, vol.~34, no.~3, pp. 178--185, 2020.

\bibitem{TSE_Highway}
T.~Seo, A.~M. Bayen, T.~Kusakabe, and Y.~Asakura, ``Traffic state estimation on
  highway: A comprehensive survey,'' \emph{Annual Reviews in Control}, vol.~43,
  pp. 128--151, 2017.

\bibitem{GraphSAGE}
J.~Liu, G.~P. Ong, and X.~Chen, ``{GraphSAGE}-based traffic speed forecasting
  for segment network with sparse data,'' \emph{IEEE Transactions on
  Intelligent Transportation Systems}, vol.~23, no.~3, pp. 1755--1766, 2022.

\bibitem{URLLC}
J.~Xue, K.~Yu, T.~Zhang, H.~Zhou, L.~Zhao, and X.~Shen, ``Cooperative deep
  reinforcement learning enabled power allocation for packet duplication
  {URLLC} in multi-connectivity vehicular networks,'' \emph{IEEE Transactions
  on Mobile Computing}, vol.~23, no.~8, pp. 8143--8157, 2024.

\bibitem{SMC_survey}
S.~Zhao, G.~Qi, T.~He, J.~Chen, Z.~Liu, and K.~Wei, ``A survey of sparse mobile
  crowdsensing: Developments and opportunities,'' \emph{IEEE Open Journal of
  the Computer Society}, vol.~3, pp. 73--85, 2022.

\bibitem{TGASA}
J.~Xue, Y.~Xu, W.~Wu, T.~Zhang, Q.~Shen, H.~Zhou, and W.~Zhuang, ``Sparse
  mobile crowdsensing for cost-effective traffic state estimation with
  spatio–temporal transformer graph neural network,'' \emph{IEEE Internet of
  Things Journal}, vol.~11, no.~9, pp. 16\,227--16\,242, 2024.

\bibitem{NC_Traffic_Jams}
M.~Saberi, H.~Hamedmoghadam, M.~Ashfaq, S.~A. Hosseini, Z.~Gu, S.~Shafiei,
  D.~J. Nair, V.~Dixit, L.~Gardner, S.~T. Waller \emph{et~al.}, ``A simple
  contagion process describes spreading of traffic jams in urban networks,''
  \emph{Nature Communications}, vol.~11, no.~1, pp. 1--9, 2020.

\bibitem{TGCN}
L.~Zhao, Y.~Song, C.~Zhang, Y.~Liu, P.~Wang, T.~Lin, M.~Deng, and H.~Li,
  ``{T-GCN}: A temporal graph convolutional network for traffic prediction,''
  \emph{IEEE Transactions on Intelligent Transportation Systems}, vol.~21,
  no.~9, pp. 3848--3858, 2020.

\bibitem{CurbGAN}
Y.~Zhang, Y.~Li, X.~Zhou, X.~Kong, and J.~Luo, ``Curb-{GAN}: Conditional urban
  traffic estimation through spatio-temporal generative adversarial networks,''
  in \emph{Proceedings of the ACM SIGKDD International Conference on Knowledge
  Discovery \& Data Mining}, 2020, pp. 842--852.

\bibitem{FillMissing_Niyato}
P.~Li, H.~Zhang, Y.~Wu, L.~Qian, R.~Yu, D.~Niyato, and X.~Shen, ``Filling the
  missing: Exploring generative {AI} for enhanced federated learning over
  heterogeneous mobile edge devices,'' \emph{IEEE Transactions on Mobile
  Computing}, 2024.

\bibitem{STGAISurvey}
Q.~Zhang, H.~Wang, C.~Long, L.~Su, X.~He, J.~Chang, T.~Wu, H.~Yin, S.-M. Yiu,
  Q.~Tian \emph{et~al.}, ``A survey of generative techniques for
  spatial-temporal data mining,'' \emph{arXiv preprint arXiv:2405.09592}, 2024.

\bibitem{AIGC_Niyato}
M.~Xu, H.~Du, D.~Niyato, J.~Kang, Z.~Xiong, S.~Mao, Z.~Han, A.~Jamalipour,
  D.~I. Kim, X.~Shen, V.~C.~M. Leung, and H.~V. Poor, ``Unleashing the power of
  edge-cloud generative {AI} in mobile networks: A survey of {AIGC} services,''
  \emph{IEEE Communications Surveys \& Tutorials}, vol.~26, no.~2, pp.
  1127--1170, 2024.

\bibitem{TSE_GPS}
A.~Abdelraouf, M.~Abdel-Aty, and N.~Mahmoud, ``Sequence-to-sequence recurrent
  graph convolutional networks for traffic estimation and prediction using
  connected probe vehicle data,'' \emph{IEEE Transactions on Intelligent
  Transportation Systems}, vol.~24, no.~1, pp. 1395--1405, 2023.

\bibitem{RAN_survey}
J.~Chen, X.~Liang, J.~Xue, Y.~Sun, H.~Zhou, and X.~Shen, ``Evolution of {RAN}
  architectures towards {6G}: Motivation, development, and enabling
  technologies,'' \emph{IEEE Communications Surveys \& Tutorials}, 2024.

\bibitem{ST_DataMining}
S.~Wang, J.~Cao, and P.~S. Yu, ``Deep learning for spatio-temporal data mining:
  A survey,'' \emph{IEEE Transactions on Knowledge and Data Engineering},
  vol.~34, no.~8, pp. 3681--3700, 2022.

\end{thebibliography}

\end{document}